\pdfoutput=1

\documentclass[11pt]{article}
\usepackage{babel}
\usepackage{graphics}
\usepackage{adjustbox}
\usepackage{float}
\usepackage{multirow}
\usepackage{longtable}
\usepackage{multicol}
\usepackage{xtab}
\usepackage{geometry}
\usepackage[dvipsnames]{xcolor}
\usepackage{soul}
\usepackage{acl}

\usepackage{times}
\usepackage{latexsym}

\usepackage[T1]{fontenc}

\usepackage[utf8]{inputenc}

\usepackage{microtype}

\title{Evaluating end-to-end entity linking on domain-specific knowledge bases: \\
Learning about ancient technologies from museum collections}

\author{\normalsize Sebastian Cadavid-Sanchez \\
  \normalsize Sciences Po \\
  \And
  \normalsize Khalil Kacem \\
  \normalsize Sciences Po \\
  \And
  \normalsize Rafael Aparecido Martins Frade \\
  \normalsize Sciences Po\\
  \AND
  \And
  \hspace*{-3cm}\normalsize Johannes Boehm \\
  \hspace*{-3cm}\normalsize Sciences Po\\
  \And
  \hspace*{-3cm}\normalsize Thomas Chaney \\
  \hspace*{-3cm}\normalsize USC \& Sciences Po\\
  \And
  \hspace*{-3cm}\normalsize Danial Lashkari \\
  \hspace*{-3cm}\normalsize Boston College\\
  \And
  \hspace*{-3cm}\normalsize Daniel Simig\thanks{\hspace{0.1cm} Correspondence to: simigd@gmail.com}\\
 }

\begin{document}
    \maketitle
\begin{abstract}
To study social, economic, and historical questions, researchers in the social sciences and humanities have started to use increasingly large unstructured textual datasets. While recent advances in NLP provide many tools to efficiently process such data, most existing approaches rely on generic solutions whose performance and suitability for domain-specific tasks is not well understood. This work presents an attempt to bridge this domain gap by exploring the use of modern Entity Linking approaches for the enrichment of museum collection data. We collect a dataset comprising of more than 1700 texts annotated with 7,510 mention-entity pairs, evaluate some off-the-shelf solutions in detail using this dataset and finally fine-tune a recent end-to-end EL model on this data. We show that our fine-tuned model significantly outperforms other approaches currently available in this domain and present a proof-of-concept use case of this model. We release our dataset and our best model.

\end{abstract}
\section{Introduction}

   
Digital technologies are enabling social scientists to easily access large quantities of textual information. These large textual datasets are also increasingly used by empirical researchers as a substitute for traditional data in numerical form \cite{gentzkow2019text}. The main challenge in doing so lies in the way that textual data is coded into numerical form. Traditionally researchers relied on experts performing the coding procedure manually, often at great expense.\footnote{As an example, the Human Relations Area Files \cite{murdock2008outline} is a curated database of text paragraphs from the cultural anthropological literature, where experts have classified each paragraph by culture and subject. The project was started in the 1940’s, and is still ongoing.} While in the past years several Entity Linking (EL) systems have been developed \cite{kolitsas2018end, gillick2019learning, wu2019zero, li2020efficient} that could possibly provide much better solutions for social scientists, these models are often developed and evaluated on generic knowledge bases (e.g. Wikidata) and datasets that are not representative of the challenges social scientists face. 

    

This works aims to enable researchers in this field to better leverage the advances in NLP. We first present a dataset we collected for the purpose of finetuning and evaluating end-to-end entity linking systems on descriptions of various museum artifacts, then use this dataset to explore the performance of different entity linking solutions at tagging and linking entities to a domain-specific knowledge base from the digital humanities. Our Knowledge Base is the \href{https://www.getty.edu/research/tools/vocabularies/aat}{Getty Arts and Architecture Thesaurus}, a controlled vocabulary describing terms associated with cultural heritage objects (predominantly materials, processes, objects, and attributes). We find that our preferred model is able to greatly outperform naive baselines currently available to researchers in the Social Sciences field.


We believe that our work contributes both to the NLP community by evaluating the generalizability of entity linking methods to different knowledge bases, as well as to the social sciences by evaluating the effectiveness of modern end-to-end entity linking methods for nuanced classification problems. We also contribute to the digital humanities and information sciences, where researchers and practitioners frequently work with controlled vocabularies and face classification problems, but are reluctant to use automated techniques like machine learning \cite{coeckelbergs2021machine} and often prefer rule-based methods or crowd-sourcing \cite{hall202113, ridge2014crowdsourcing}. We offer an off-the-shelf solution that can be used and further developed by practitioners.
\section{The ARTIFACT Dataset}

\begin{figure*}[!h]
\begin{centering}
\includegraphics[scale=0.6]{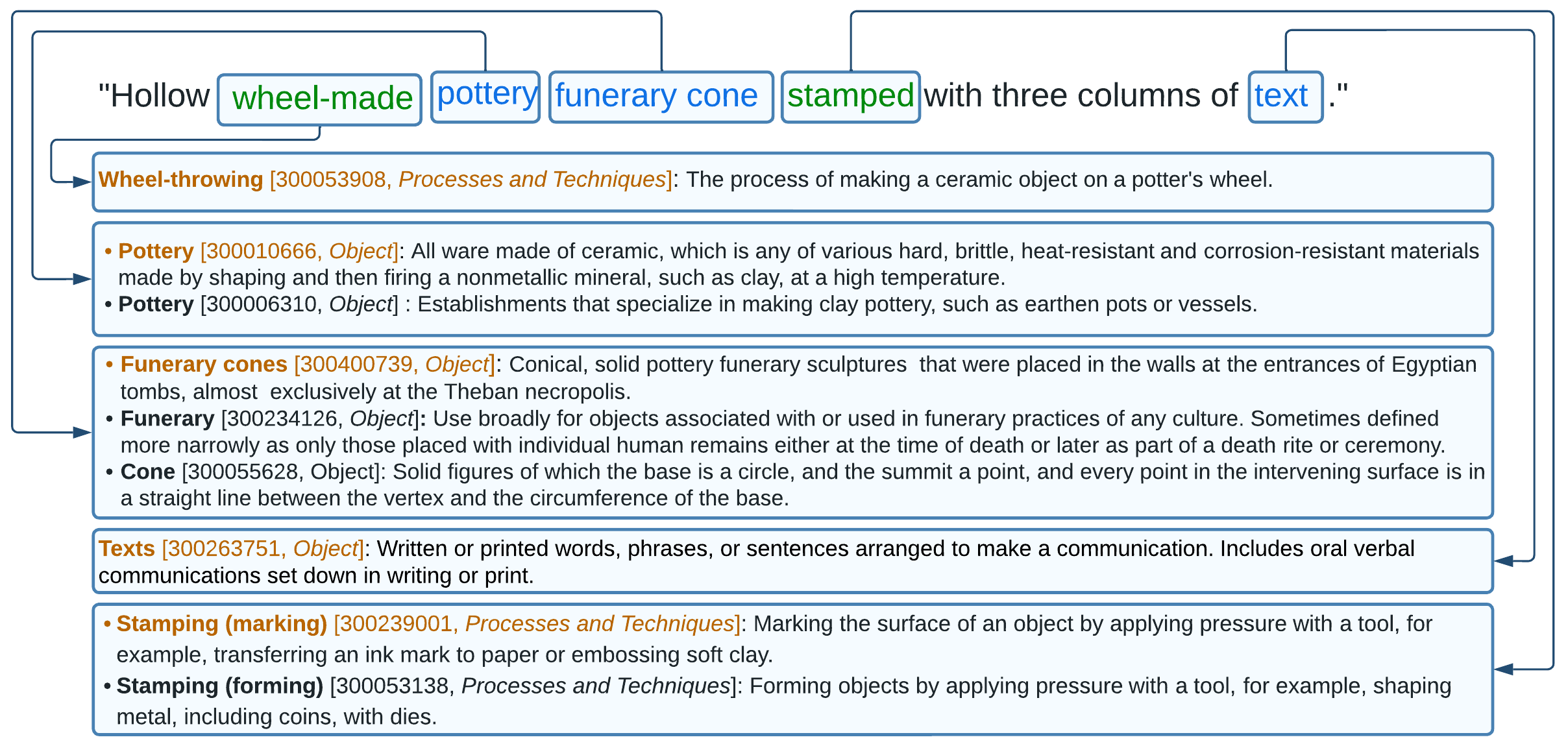}
\par\end{centering}
\caption{Example of end-to-end entity linkage of a museum item description. Each box contains recognized entities (Mention Detection). The green text represents the Process and Techniques facet, and the blue text represents the Object facet. The light-blue boxes include the most relevant candidates for entity linking (Candidate Generation), and the gold text indicates the correct candidate (Entity Disambiguation). Each integer identifier represents the corresponding AAT ID for each candidate. 
\label{fig:end_to_end_example}}
\end{figure*}


\subsection{Sourcing Museum Labels}
\renewcommand{\arraystretch}{1.2}


To create this dataset we use textual descriptions of objects from six online museum collections\footnote{British Museum, Metropolitan Museum of Art, Williams College Museum, Cleveland Museum of Art, Penn Museum and the Smithsonian Museums. The data in this paper is part of a broader project on ancient technologies, using tens of millions of objects from dozens of museums}. Each object in these collections is annotated with 4 different text fields: title, detailed description, and free-form metadata about materials and techniques. Table \ref{tab:ex_description_types} provides examples.

In some cases, different types of text fields may be repetitive in terms of entity content and contain texts of different complexity. In order to efficiently use our resources and avoid redundancy, we sample from different types of text fields independently.

\begin{table}[h]
\begin{centering}
\begin{tabular}{p{1.7cm}p{5.3cm}}
\hline
\hline
\textbf{\small{}Type} & \textbf{\small{}Example}\tabularnewline
\hline 
{\small{}Title} & {\small{}''Fragment of Carved Wood Panel with Two Birds''}\tabularnewline
{\small{}Description} & {\small{}''Fragment. With part of figure of ghulám, sleeve.
Made of black, grey-brown, blue painted plaste''}\tabularnewline
{\small{}Material} & {\small{}''Gilt-copper alloy, rock crystal''}\tabularnewline
{\small{}Technique} & {\small{}''Plain weave, tapestry, dyed''}\tabularnewline
\hline
\end{tabular} 
\caption{Examples for each object description type in our digital museum data}\label{tab:ex_description_types}
\end{centering}
\end{table}

To evaluate the performance of different Entity Linking techniques, we take a small sample from this dataset and manually annotate it with EL labels. Since some materials and objects appear much more frequently than others, we increase the variability of labels in our sample by implementing a stratified sampling process based on the registered date for each object.  We provide more details on the sampling process in Appendix \ref{subsec:app_data_sampling}.

\subsection{Using a domain-specific knowledge base\label{subsec:domain_spec}}
Unlike most Entity Linking applications that use Wikipedia as their knowledge base, our dataset links entities to the \emph{Getty Arts and Architecture Thesaurus} (AAT),\footnote{\url{https://www.getty.edu/research/tools/vocabularies/aat/about.html}} an expert-curated taxonomy of more than 50,000 concepts related to arts, architecture and visual cultural heritage. All concepts are structured hierarchically in eight facets (subtrees) \cite{AboutAAT}. Each entry has a definition and belongs to a rank in the hierarchy structure. We focus on four facets of the taxonomy, listed in Table\,\ref{tab:ex_hierachy_types}.

\begin{table}[h]
\begin{centering}
\begin{tabular}[h]{p{2.5cm}p{4.5cm}}
\hline
\hline 
\textbf{\small{}Hierarchy} & \textbf{\small{}Examples}\tabularnewline
\hline 
{\small{}Objects} & {\small{}Knifes, vessels, amphorae, neck, earthenware, handle.}\tabularnewline
{\small{}Materials} & {\small{}Ceramic, gold, clay, wood, calcite, diamond, glass.}\tabularnewline
{\small{}Styles and Periods} & {\small{}Periods Peruvian, ancient chinese, greek, gothic,
inca.}\tabularnewline
{\small{}Activities} & {\small{}Machine sewing, sealing, gilding, hammering.}\tabularnewline
\hline

\end{tabular} 
\caption{Examples for analyzed hierarchies of the taxonomy \label{tab:ex_hierachy_types}}
\end{centering}
\end{table}

AAT is different from Wikipedia in many ways:

\begin{itemize}
\item AAT focuses on a small subset of domains covered by Wikipedia but provides a much more granular coverage of entities in those domains. For example, a search in Wikipedia for the word "hammer" returns a few different results, while AAT contains more than one hundred different types of hammers. The main differences between types lie in distinctive characteristics, such as specific use, material, period of manufacture, etc.
\item Unlike in the case of Wikipedia that has a hyperlink graph, in our setting one cannot leverage pre-computed mention tables to aid candidate generation.
\item Descriptions in AAT are much shorter than in Wikipedia--their purpose is merely to provide a definition of the entity in question. On average, AAT descriptions are two sentences long. 
\item In our dataset we focus solely on non-named entities and ignore any named entities - in contrast with a large body of literature focused on named entities in Wikipedia.
\end{itemize}

To the best of our knowledge, all prior approaches used by scholars in the digital humanities to link text to AAT entities involve manual annotation, in some cases supported by string and keyword matching \citep{OpenRefineRec, MatchGettyAAT}.

\renewcommand{\arraystretch}{1}

\subsection{Annotation Process}

To produce manually-annotated data, sampled strings were distributed evenly among 4 annotators. Instruction lessons were conducted to show the annotators gold examples and orient them to use an in-house tool to produce and store annotations.

Throughout the annotation process, a quality control exercise was repeatedly carried out. In this the 4 annotators received an identical, small batch of data, annotated them independently, then discussed their results together and recorded the consolidated labels. This process ensured that labels from different raters are consistent, but also resulted in a smaller dataset (552 annotated mention-entity pairs) that we use as a high quality test set for our evaluations. Table \ref{tab:human_metrics} shows performance statistics of individual raters when compared to the commonly agreed-upon ground truth. Notably, we found that there exists some amount of unavoidable ambiguity in the Entity Linking task due to the limited information available in descriptions - even our best performing rater could only produce and end-to-end F1 score of 86.8 when compared to the consensus labels.


\begin{table}[]
\begin{centering}
\resizebox{\columnwidth}{!}{
\begin{tabular}{llcccc}
\hline\hline 
 & Metric (\%) & Upper bound & Median & Lower bound \tabularnewline
\hline 
\multirow{3}{*}{MD} & Precision & 94.8 & 92.4 & 88.9 \tabularnewline
 & Recall & 90.6 & 80.9 & 71.4  \tabularnewline
 & F1 & 92.2 & 86.5 & 79.2\tabularnewline
\hline 
ED & Precision & 88 & 81 & 71\tabularnewline
\hline 
\multirow{3}{*}{End-to-end} & Precision & 88.3 & 85.6 & 82.9\tabularnewline
 & Recall & 85.3 & 75.8 & 67.9\tabularnewline
 & F1 & 86.8 & 80.4 & 74.7\tabularnewline
\hline 
 &  &  &  &  & \tabularnewline
\end{tabular}}
\par\end{centering}
\caption{Performance statistics on the jointly labeled data. We consider the commonly agreed label as the gold label and measure strong-matching performance of individual raters' agains that. \label{tab:human_metrics}}
\end{table}

The final, single-annotated set comprises of 1,728 strings annotated with a total of 7,510 entity labels. This annotation process (including discussion sessions) took approximately 300 hours of human labor. Given the difficulty of the Entity Linking task, it was initially costly to standardize criteria between annotators for disambiguating between different entities in different contexts. 

We use the single-annotated dataset to finetune models for our domain. We randomly split this dataset into training (90\%) and validation (10\%) sets. Main statistics of these data sets are presented in Table \ref{tab:data_comp_type} and additional details are provided in Appendix \ref{subsec:annotations_composition}. Our dataset called ARTIFACT\footnote{Automated Resource for Technology and Information Finding in Artifacts CollecTions.} is available at: \href{https://github.com/LongRunGrowth/ARTIFACT-Dataset}{https://github.com/LongRunGrowth/ARTIFACT-Dataset}.

\subsection{Challenges of Entity Linking in digital museum data}


We highlight several Entity Linking challenges presented by our dataset, using the example in Figure \ref{fig:end_to_end_example}. We believe these properties make ARTIFACT a suitable dataset to finetune and evaluate generic EL systems on this specific domain.


\paragraph{Context-dependent disambiguation}
In 43.5\% of the cases, matching semantic context from the text and the entity definition is crucial to make the correct selection, as multiple entities with similar names could be a possible match based on the surface form.\footnote{To calculate this metric, we sampled over 30\% of errors generated by our string-matching baseline (described in section \ref{sec:baselines}) and classified them into context-dependent and non-context-dependent errors. Using this metric, we approximate the proportion of context-dependent disambiguation cases over our full annotations set.} In Figure \ref{fig:end_to_end_example}, the mention "pottery" has two AAT candidates with identical names; however, their descriptions vary substantially. Using the context suggests the text does not refer to a workshop where pottery is made, but to a pottery object. A similar case applies to the word "stamped" in the example.

\paragraph{Surface form diversity.} 9.5\% of the labels in our dataset have a significantly different entity name and mention text. This is a failure case for the currently prevalent fuzzy-matching based methods and currently requires an annotator to do a detailed search or know the entity in advance. This is illustrated in Figure \ref{fig:end_to_end_example} for the "wheel-made" mention:  usually, this term refers to pottery made using a wheel. However,  "wheel-made" does not match exactly with any entity in our knowledge base, but a similar expression with the same meaning exists, "wheel-throwing". Using a simpler linking algorithm, such as string matching, this type of mention could be mapped to the incorrect entity "wheel making", which corresponds to "the art of making wheels".

\paragraph{Overlapping Mentions} This is exemplified in Figure \ref{fig:end_to_end_example} by the mention "funerary cone". Although there is an entity that matches exactly the full mention, an annotator may tag "funerary" and "cone" and link them to separate entities.

\paragraph{Incomplete Taxonomy} We note that there were cases where even though a referenced entity belonged to one of the analyzed facets, it was not currently included in AAT and thus could not have been included in our dataset. The incompleteness of taxonomies for high-cardinality classification tasks is a known problem \citep{simig2022open}, and addressing it is beyond the scope of this work.

\section{Baseline Systems\label{sec:baselines}}

We provide and evaluate a set of baseline EL solutions along with the dataset, covering a wide range of system complexities. We first test model-free approaches currently available, as well as off-the-shelf machine learning based systems. Finally, we adapt a more recent model originally trained on Wikipedia (ELQ, \citealt{li2020efficient}) to our domain.

\subsection{Parameter-free Baselines}

\textbf{Naive String Matching} We build a pretrained SpaCy \cite{matthew_honnibal_2019_3358113} pipeline that provides an end-to-end framework by integrating an i) \textquotedbl Entity-Recognizer\textquotedbl\footnote{\href{https://spacy.io/api/entityrecognizer}{https://spacy.io/api/entityrecognizer}} model to predict non-overlapping mention boundaries using fuzzy matching to search spans from entities in the knowledge base and an ii) \textquotedbl Entity-Ruler\textquotedbl\footnote{\href{https://spacy.io/api/entityruler}{https://spacy.io/api/entityruler}} model to perform Candidate Generation and Entity Disambiguation tasks. The model performs random candidate disambiguation when there are similar matches between candidates. Using the SpaCy-pipelines platform allows users wide accessibility and fast usage, and to the best of our knowledge, is the most accessible tool for entity-linking tasks on museum data before our work. 

\textbf{Memorization} We establish a rule-based entity disambiguation baseline in which we link a mention in the test set to the entity most commonly associated with this mention in the train set. For example, the string "buff" could be tagged to two different entities (a tool or a pottery style). This ruled-based baseline would tag all occurrences of "buff" with the most common entity in the train set. We expect systems that truly utilize context (as opposed to just overfitting on the seen mention-entity pairs) to outperform this baseline significantly.

\subsection{Model-Based Disambiguation}

\textbf{SBERT} \cite{reimers-2020-multilingual-sentence-bert} computes text embeddings that can be used to identify similar sentences. As a simple baseline, we use our AAT Knowledge Base to create embeddings for each entity description and similarly embed the text in which the mention appears. We choose the entity with the highest cosine similarity.

\textbf{GENRE} \cite{decao2021autoregressive} is a sequence-to-sequence model that uses BART \cite{lewis2019bart} to link tagged mentions to entities from Wikipedia using constrained beam search to generate valid identifiers.

In order to link the output of GENRE to AAT entities, we build a database that maps Wikipedia IDs and labels with their corresponding AAT when it exists. When the Wikipedia label is not specific enough and corresponds to different AATs (as in the example provided in subsection \ref{subsec:domain_spec}), a candidate is randomly selected. We expect this to be a weak baseline that highlights the limitations of systems that rely on Wikipedia as their knowledge base.

\subsection{An End-to-End Solution}

\textbf{ELQ} \cite{li2020efficient} is an entity linking system that performs both mention detection and disambiguation, originally developed to help question-answering models. ELQ detects mentions and computes mention representations in a single pass of inference, retrieves entity candidates using a K-Nearest Neighbor (KNN) search and chooses the best candidate based on embedding similarity. We use this model as our strongest baseline as we found it to provide the best trade-off between low system complexity and good performance on standard standard EL benchmarks.

We concatenate names and descriptions of AAT entities to form the text to be encoded by ELQ's bi-encoder. We first evaluate an off-the-shelf version of the ELQ model provided by \citet*{li2020efficient}, then also finetune this on our training set in order to adopt it to our domain specific texts and entities. We run all training exercises for 500 epochs using a batch size of 256 on a single NVIDIA A100 GPU. We use AdamW optimizer with learning rate 1e-5 and linear decay.

\section{Evaluation of Baseline Systems\label{sec:Evaluation}}

\subsection{Evaluation Protocol}\label{subsec:metrics}

We separate evaluation into three sections: Mention Detection, Entity Disambiguation, and End-to-End performance. To evaluate mention detection, we use strong matching: an annotation is correct if its boundaries match the ground-truth mention boundaries in the text exactly. When evaluating entity disambiguation, an annotation is correct if the linked entity ID for a given mention precisely matches the AAT ID in the ground-truth annotation. An End-to-End annotation is considered correct only if it matches both the mention boundaries and AAT ID of the ground-truth annotation exactly.

In order to compare Entity Disambiguation performance among a wide range of disambiguation solutions (Table \ref{tab:linking_results}), we select as input the correctly tagged mentions from the ELQ model. For this experiment we further divide tagged items into (i) \textit{Seen} tags that are present in annotations of both the training set and the test set (88.94 \%) and (ii) \textit{Unseen} Tags that did not appear in the training set (11.06\%). The latter set - albeit small - allows us to evaluate the generalization capacity of our models when considering unobserved tags, which could appear considerably in our application case. 

\subsection{Results}\label{subsec:results}

The results of evaluation for Mention Detection, Entity Disambiguation and End-to-End Entity Linking on our held out, high quality test set are presented in Tables \ref{tab:md_results}, \ref{tab:linking_results} and \ref{end_to_end_algorithms} respectively.

\begin{table}[H]
\resizebox{\columnwidth}{!}{
\begin{tabular}{lcccc}
\hline\hline 
 &  & \multicolumn{3}{c}{\textbf{Metric (\%)}}\tabularnewline
\cline{3-5} 
\textbf{Annotation-system} & \textbf{Model version} & \textbf{Precision} & \textbf{Recall} & \textbf{F1}\tabularnewline
\hline 
String Matching & Off-the-shelf & 57.6 & 42.9 & 49.1\tabularnewline
ELQ & Off-the-shelf & 62.9 & 44.2 & 51.9\tabularnewline
ELQ & Fine-tunned & \textbf{85.4} & \textbf{81} & \textbf{83.1}\tabularnewline
\hline 
\end{tabular}
}
\caption{Performance for annotation systems able to perform Mention Detection.\label{tab:md_results}}
\end{table}

\begin{table}[H]
\resizebox{\columnwidth}{!}{
\begin{tabular}{lcccc}
\hline\hline 
 & & \multicolumn{3}{c}{\textbf{\% Precision}}\tabularnewline
\cline{3-5}
\textbf{Annotation-system} & \textbf{Model version} & \textbf{All tags} & \textbf{Unseen} &\textbf{Seen}\tabularnewline
 \hline 
String-Matching & Off-the-shelf & 45.5 & 49.8 & 44.5\tabularnewline
Memorization baseline & - & 78.6 & 0 & 87.4\tabularnewline
SBERT & Off-the-shelf & 64.6 & \textbf{74.6} & 62.7\tabularnewline
GENRE & Off-the-shelf & 30.5 & 43.1 & 28.7\tabularnewline
ELQ & Off-the-shelf & 39.9 & 43.4 & 8.2\tabularnewline
ELQ & Fine-tuned & \textbf{88.7} & 68.4 & \textbf{90.2}\tabularnewline
\hline 
\end{tabular}}
\caption{Performance for annotation systems able to perform Entity Linking. \label{tab:linking_results}}
\end{table}

\begin{table}[H]
\resizebox{\columnwidth}{!}{
\begin{tabular}{lcccc}
\hline \hline 
 &  & \multicolumn{3}{c}{\textbf{Metric (\%)}}\tabularnewline
\cline{3-5}
\textbf{Annotation-system} & \textbf{Model version} & \textbf{Precision} & \textbf{Recall} & \textbf{F1}\tabularnewline
\hline
String Matching & Off-the-shelf & 59.6 & 39.5 & 47.5\tabularnewline
ELQ & Off-the-shelf & 57.7 & 42.2 & 48.7\tabularnewline
ELQ & Fine-tunned & \textbf{78.5} & \textbf{74.6} & \textbf{76.9}\tabularnewline
\hline 
\end{tabular}}
\caption{Performance for annotation systems able to perform entity linking End-to-End.\label{end_to_end_algorithms}}
\end{table}

Even a state-of-the-art model such ELQ performs only on par with simple baselines when not adapted to the domain of our texts and the custom knowledge base, AAT. However, fine-tuning ELQ even on our relatively small training dataset results in a step change in the model's performance. Notably, Entity Disambiguation performance on the \textit{Unseen} set also improves, demonstrating that the ELQ model learns to generalize from the training set we provide. Interestingly, the off-the-shelf SBERT solution for disambiguation outperforms finetuned ELQ when linking unseen AATs, indicating that there is still room for improvement for future end-to-end models. Despite that, the fine-tuned ELQ model reaches the lower-bound of human performance metrics for Mention Detection and End-to-end tasks.

\section{An Example Use Case}

\begin{figure*}[h!]
\begin{centering}
\includegraphics[scale=0.38]{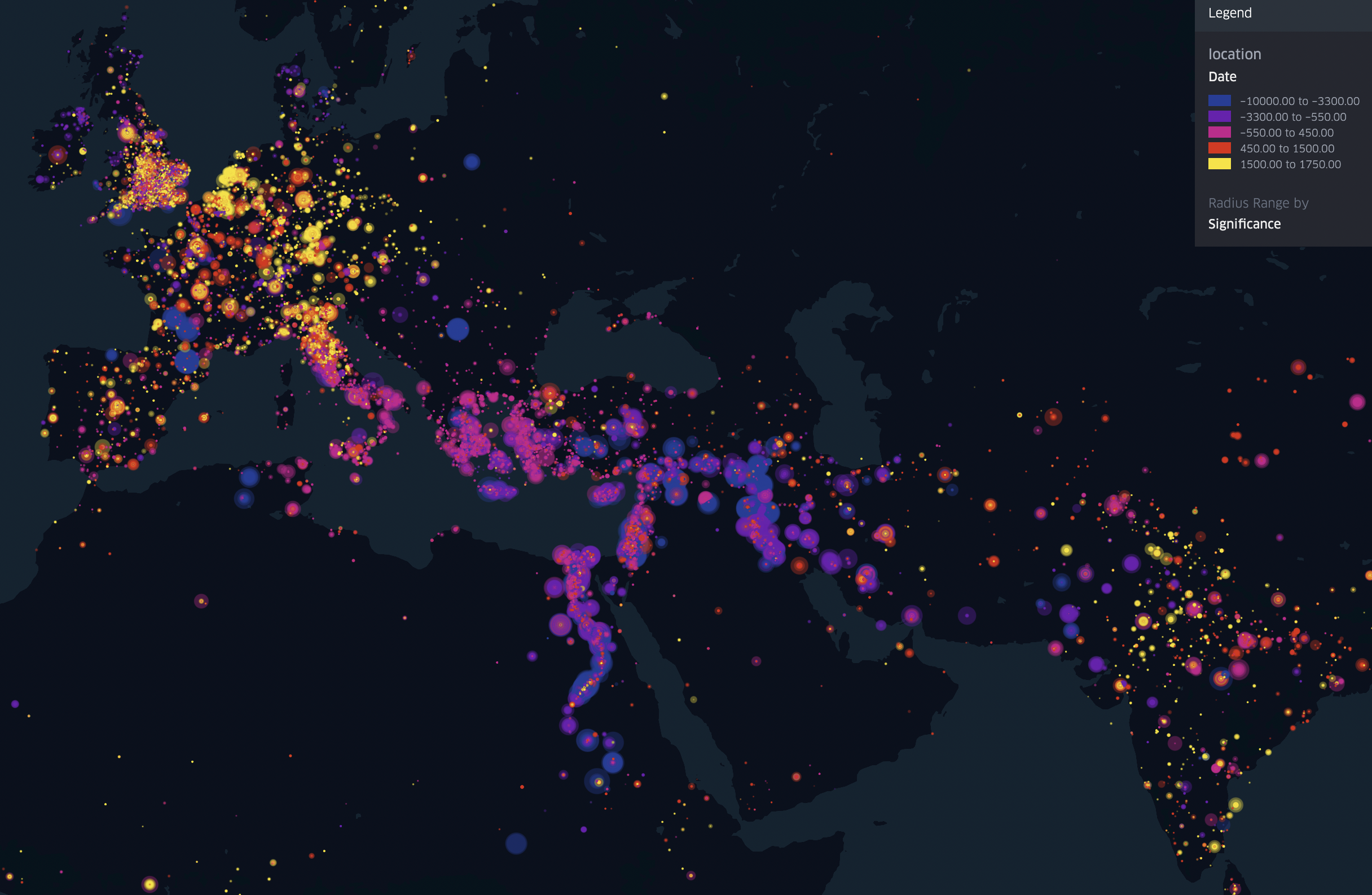}
\par\end{centering}
\caption{This figure maps 2.7 million objects from 26 English-language museum collection, where we classify objects into AAT technology categories using our End-to-End Entity Linking system (ELQ fine-tuned). A circle's color corresponds to the date when objects are made (blue: Neolithic; purple: Bronze and Iron Ages; pink: Classical Antiquity; red: Middle Ages; yellow: Renaissance). A more opaque circle corresponds to more objects in the same location. The radius of each circle corresponds to a measure of the `technological significance' of an object, constructed using our End-to-End classification. For object $o$ that belongs to AAT technology category\,$c$: $significance_{o}^{c}=\ln(Number_{c})/Rank_{o}^{c}$, where $Number_{c}$ is the total number of objects in technology category $c$, and $Rank_{o}^{c}$ is the temporal rank of object $o$ among all objects in technology category $c$ ($Rank_{o}^{c}=1$ if object $o$ is the earliest in technology category $c$, $Rank_{o}^{c}=2$ if it is the second earliest, etc).
\label{fig:proof_of_concept_map}}
\end{figure*}

Figure \ref{fig:proof_of_concept_map} shows, as a proof-of-concept example, that entity linking can a powerful tool to study the invention and diffusion of technologies over time and space. We apply our End-to-End entity linking system to 2.7 million objects from 26 English-language museum collections. We define an object as `technologically significant' (large radius) if it belongs to an important technology class (a class with many objects) and if it is a pioneer within its technology class (early discoveries). 

The map allows to visualize the process of technological diffusion over more than ten millennia. Neolithic sites with significant technologies (large blue circles) appear along the Nile and Indus rivers, Southern Turkey and the fertile crescent, as well as in Southwest France. Other less significant Neolithic sites are scattered over Europe, the Middle East, and South Asia (small blue circles). Over the Bronze and Iron Ages (purple), significant technologies are concentrated along the Nile, the Indus, and the fertile crescent. Classical Antiquity (pink circles) sees a shift of significant technologies to Greece, the Near East, and the Southern Italian peninsula. In the Middle Ages and the Renaissance (red and yellow circles), significant technologies are concentrated in Western Europe and South Asia. 

Crucially, our ability to link individual objects to AAT entities is key to identifying `significant' technologies. For instance, focusing on the Renaissance (yellow circles), while objects appear throughout Europe, the most significant discoveries (large yellow circles) are concentrated in Northern Italy, and Northern and Central Europe, as well as in important urban centers such as Paris, London, and Madrid.

\section{Related Work}


\subsection{Entity Linking for Sciences}


Entity Linking has proved to be a powerful tool in various fields of science. \citet*{d2020stem} present the STEM-ECR dataset with annotations to identify and classify scientific entities from abstracts of publications drawn from 10 STEM (Science, Technology, Engineering, and Medicine) disciplines. Applications of EL in biomedicine \cite{mohan2019medmentions, medtype2020,liu2021learning} and social network analysis \cite{derczynski2015analysis, adjali2020multimodal} have also been explored.

\subsection{NLP for Social Sciences}
\citet*{weichselbraun2022slot} introduce the CarrerCoach Dataset, which contains annotations across 75 documents related to education and labor market entities; and \citet*{tsereteli2022overview} present the SV-Ident dataset with annotations to perform survey variable identification in social science publications. \citet*{gillick2019learning} introduces an Entity Linking dataset for news article analysis. To the best of our knowledge, our dataset is the first to provide a manually annotated database intended for quantitative social sciences research using museum collections data.  

We see two broad fields of applications of our work. One is in the quantitative social sciences, where textual information is increasingly being used to construct numerical data that is being processed using statistical and econometric techniques (\citealp{gentzkow2019text}, \citealp{wilkerson2017large}, and \citealp{jensen2022language} give overviews of such approaches in economics, political science, and sociology, respectively). So far, much of the work in this domain uses unsupervised learning methods, in particular topic modelling, to reduce the dimensionality of the data (e.g. \citealp{hansen2018transparency}, \citealp{mueller2018reading}). When the categories that observations should be classified into are known ex ante, these approaches become difficult to use. That is the case in particular in economics, where there is often a consensus on the use of certain industry classifications or patent codes. Related to our application, the use of machine learning techniques is particularly promising in the field of economic history, where much of the original data is in textual form (see, e.g. \citealp{feigenbaum2016machine}, \citealp{combes2022urban}, or \citealp{shen2020large} for examples).

The second field of applications are the digital humanities, where researchers and practitioners are often confronted with large quantities of unstructured text data. Entity linking provides a way of classifying and thereby linking records from these corpora, enhancing the usefulness and discoverability of these datasets (\citealp{knoblock2017lessons}). Our work may be of particular interest to those working on cultural heritage data, where the AAT is a commonly used controlled vocabulary (see, e.g. \citealp{van2013evaluating}, or the Linked Art project, described in \citealp{delmas2020fostering}). Previous approaches in the digital humanities community either use manual entity linking (e.g. \citealp{van2013evaluating}) or use approaches based on term frequencies (e.g. \citealp{mendes2011dbpedia}). \citet{MUNNELLY2018199} describe the process of creating a domain-specific knowledge of notable Irish people and constructing an entity linker via DBpedia. \citet{van2015exploring} evaluate different web-based NEL services for the linking of museum collection metadata to Wikidata, but find recall rates of below 50\% using these services. \citet{brando:hal-01396037} present an domain-specific graph-based entity linking strategy for historical figures. Most closely related to our approach, \citet{linhares2022melhissa} train an end-to-end entity linker based on MUSE (\citealp{conneau2017word}) for linking entities in historical press articles to Wikipedia entries.



\section{Future Work}

Our evaluation was performed on an English-language corpus and knowledge base. The Getty AAT contains translations of many records, facilitating an evaluation of entity linking pipelines in other languages. We leave this to future research.  

A different and complementary approach to classify the objects in our dataset would be to make use of photos that are associated with them. While photos are not available for the entire dataset and many of the available images do not capture the objects in detail, a possible improvement would use multimodal algorithms that combine images and text descriptions to link entities.

\section*{Acknowledgements}
The authors are grateful to the Data Intelligence Institute of Paris (diiP) for financial support and to TGIR Huma-Num (CNRS) for computational infrastructure support. Chaney thanks the European Research Council for financial support (ERC grant N\textsuperscript{o}884847).

\bibliography{anthology,custom}
\bibliographystyle{acl_natbib}

\appendix

\section{Appendix}
\label{sec:appendix}

\subsection{Data Sampling\label{subsec:app_data_sampling}}

There are significant imbalances in the frequency of words and expressions in the raw data that various museums provide. A significant part of this data consists of single word descriptions, like "paintings", "coins" and "figure". In order to have more information-rich data to annotate, we sampled a different number of examples of different types of fields independently. Our ARTIFACT dataset consists of:
\begin{itemize}
  \item 50\% Descriptions,
  \item 30\% Titles,
  \item 20\% Materials and techniques.
\end{itemize}

An effect of this sampling method is that for instance the presence of the "description" field for object A in our dataset does not mean that "title" filed of the same object will also be annotated with EL labels.

Another important characteristic of the data is that, since it comes from museum collections, objects are more numerous for some time periods  than others. To have a more diversified sample, respecting the proportions mentioned above, we uniformly sampled descriptions from time windows, that is, we randomly selected the same number of strings from every two hundred years window after the year zero and from every five hundred years window before the year zero till 3000 b.c.

\subsection{Dataset  statistics\label{subsec:annotations_composition}}
Single-annotated data were generated once inter-annotator agreement rates were stable between previous annotation sessions. Additionally, for the single-annotation datasets, an additional data cleaning check was performed to filter or correct manual annotation errors related to label span detection. Table \ref{tab:data_comp_type} shows the general statistics for the composition of the dataset.

\begin{table}[h]
\begin{centering}
\resizebox{\columnwidth}{!}{
\begin{tabular}{ccccc}
\hline \hline 
\textbf{Data type} & \textbf{Annotations} & \textbf{US} & \textbf{UE} & \textbf{UM}\tabularnewline
\hline 
Single-annotated & 7,510 & 1,728 & 1,500 & 1,802\tabularnewline
Cross-annotated & 552 & 127 & 296 & 315\tabularnewline
\hline
\end{tabular}}
\par\end{centering}
\caption{Data composition by annotation type: Unique strings (US), Unique Entities
(UE), and Unique mentions (UM).\label{tab:data_comp_type}}
\end{table}

\begin{figure}[h]
\begin{centering}
\includegraphics[scale=0.60]{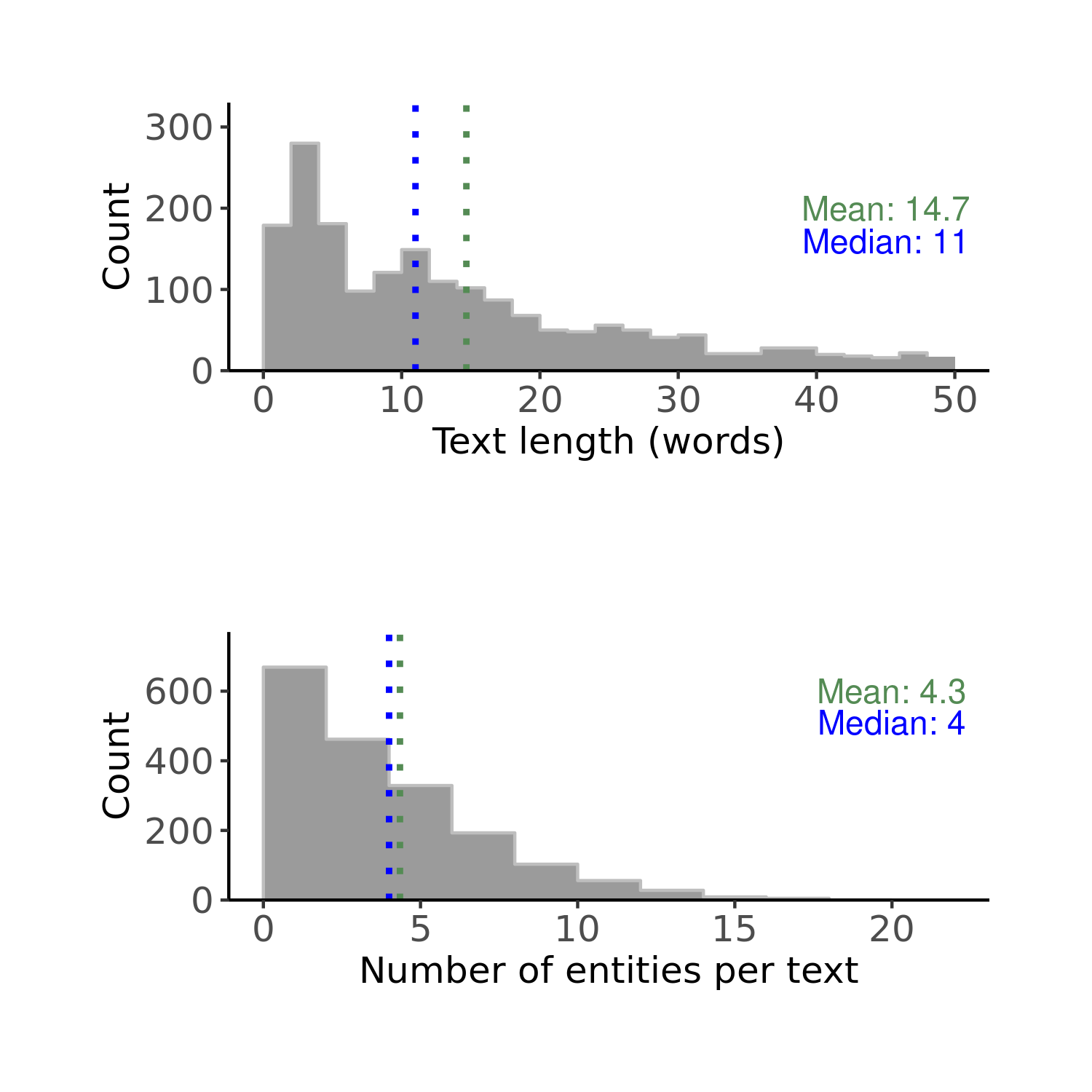}
\par\end{centering}
\caption{Top panel: Distribution of text length (words). Bottom Panel: Distribution of entities per text.\label{fig:text_dists}}
\end{figure}

Figure \ref{fig:text_dists} shows the distribution of the length of the text (number of words) for object description and the distribution of the number of entities per text for the whole dataset. On average, the descriptions in our dataset are between 14 and 15 words long and have four entities. 

Table \ref{tab:human_metrics} shows the size of the text descriptions, the number of entities per text, and the number of unique entities for training, validation, and test sets, respectively. 

\begin{table}
\begin{centering}
\resizebox{\columnwidth}{!}{
\begin{tabular}{lcccc}
\hline \hline 
 &  & \multicolumn{3}{c}{\textbf{Dataset}}\tabularnewline
\cline{3-5}
\textbf{Facet} & \textbf{Metric} & \textbf{Train} & \textbf{Validation} & \textbf{Test}\tabularnewline
\hline 
\multirow{2}{*}{Activities} & C & 1,107 & 132 & 431\tabularnewline
 & UE & 207 & 66 & 66\tabularnewline
\multirow{2}{*}{Materials} & C & 1,658 & 180 & 699\tabularnewline
 & UE & 309 & 84 & 99\tabularnewline
\multirow{2}{*}{Objects} & C & 3,830 & 433 & 1,316\tabularnewline
 & UE & 795 & 219 & 201\tabularnewline
\multirow{2}{*}{Styles \& Periods} & C & 159 & 11 & 43\tabularnewline
 & UE & 98 & 9 & 13\tabularnewline
\hline
\end{tabular}}
\par\end{centering}
\caption{Entity counts per facet (C), and unique entity counts per facet
(UE) for training, validation, and test sets.\label{tab:ec_per_data_split}}
\end{table}

The most common elements in our facets of interest are objects and materials. Table \ref{tab:ec_per_data_split} shows the 
entity counts per facet, and unique identity counts per facet for each data split. 

\begin{figure}[h]
\begin{centering}
\includegraphics[scale=0.35]{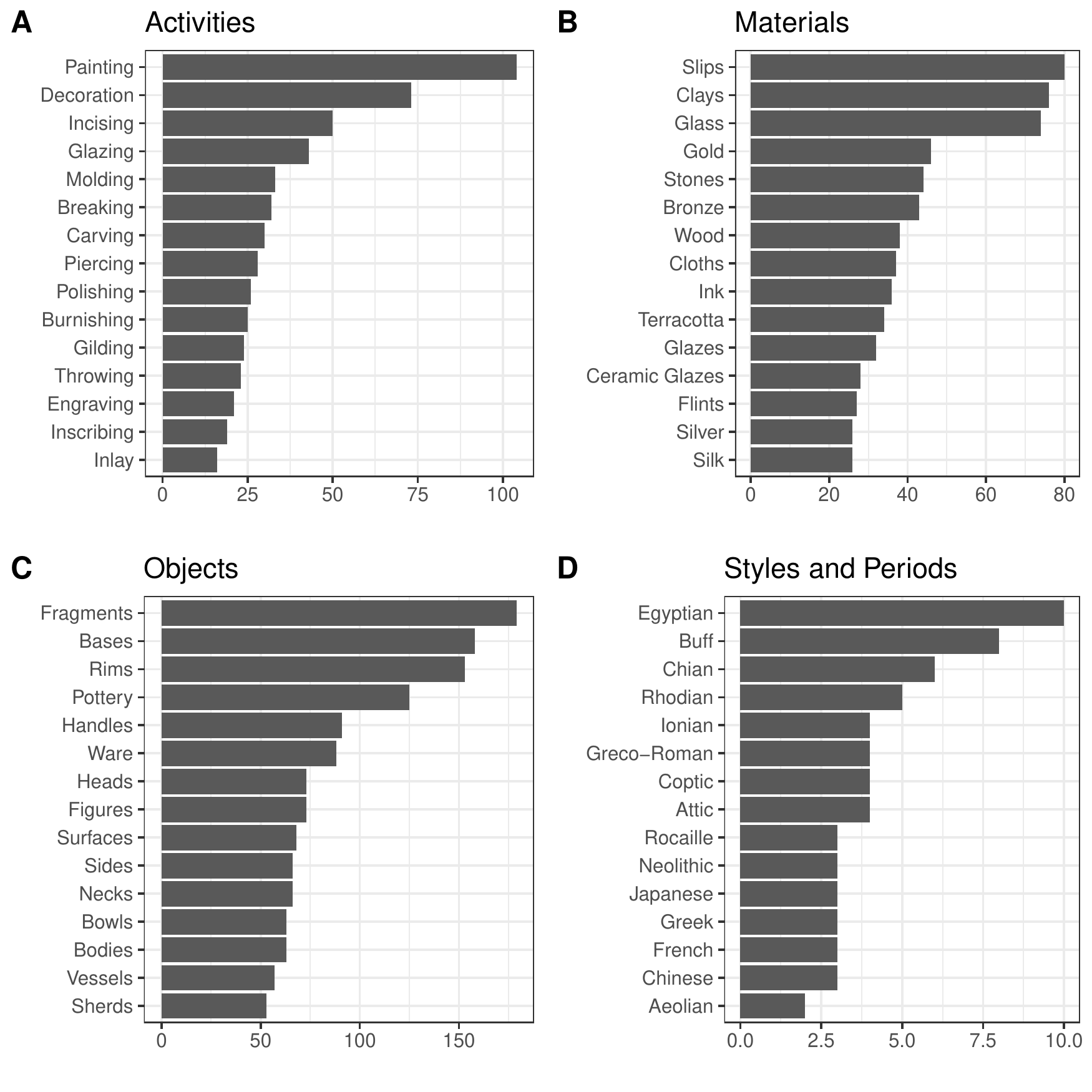}
\par\end{centering}
\caption{Top-10 entities by facets and counts for the training set}
\label{fig:top10_training}
\end{figure}

\begin{figure}[h]
\begin{centering}
\includegraphics[scale=0.35]{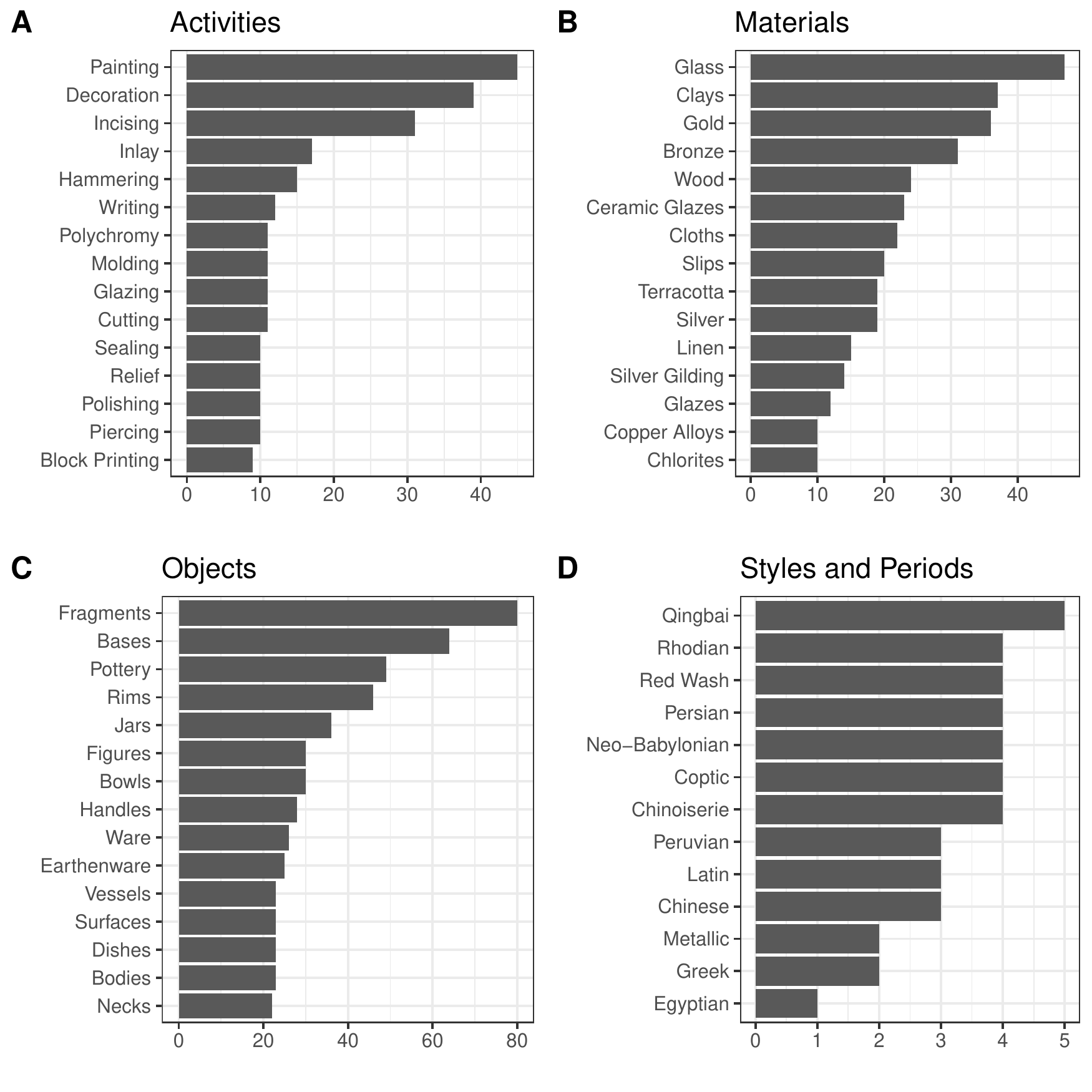}
\par\end{centering}
\caption{Top-10 entities by facets and counts for the test set}
\label{fig:top10_test}
\end{figure}

\begin{figure}[h]
\begin{centering}
\includegraphics[scale=0.35]{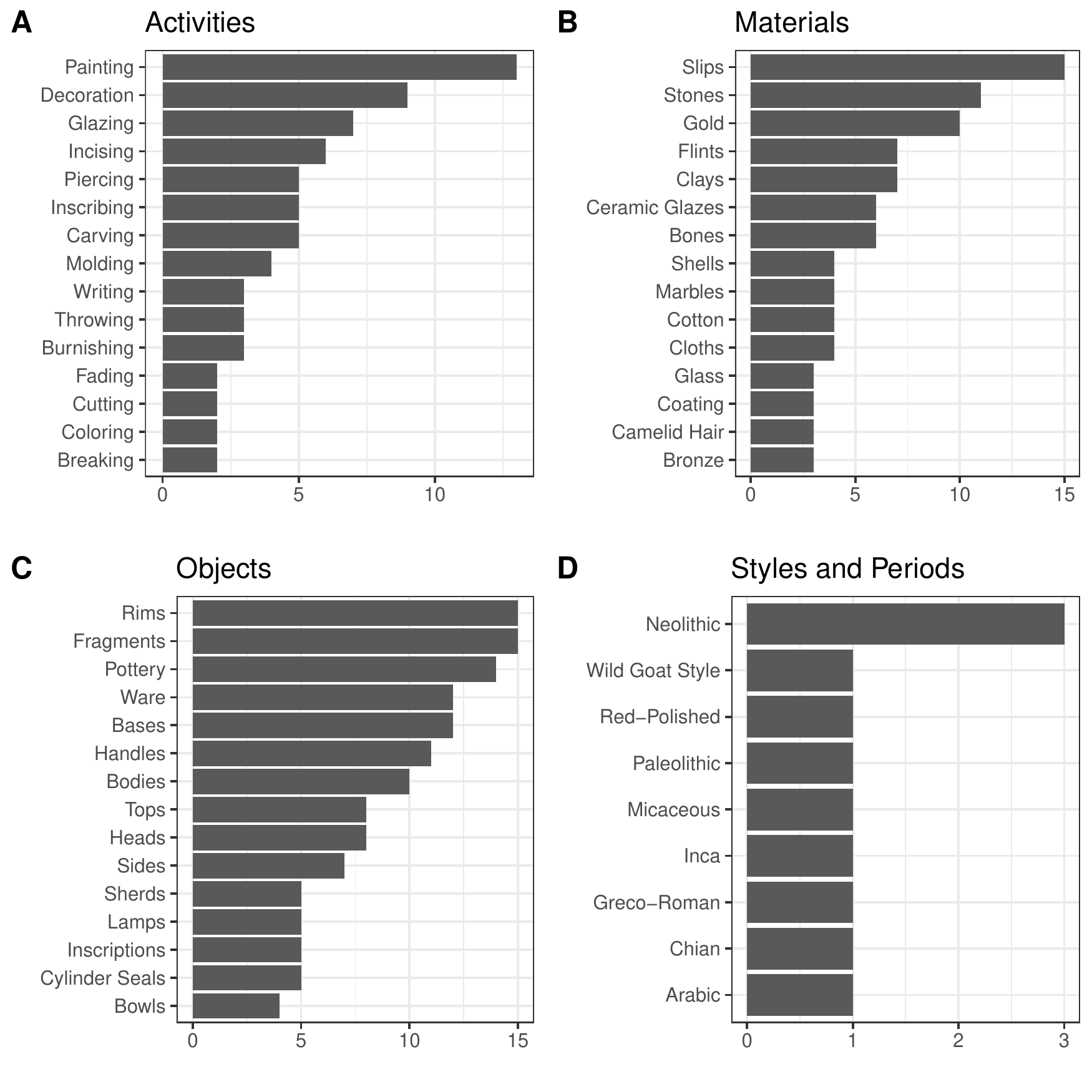}
\par\end{centering}
\caption{Top-10 entities by facets and counts for the validation set}
\label{fig:top10_val}
\end{figure}

The top-10 most frequent entities by facet are provided in figures \ref{fig:top10_training}, \ref{fig:top10_test}, and \ref{fig:top10_val}.

\end{document}